\documentclass[review]{elsarticle}

\makeatletter
\def\ps@pprintTitle{%
 \let\@oddhead\@empty
 \let\@evenhead\@empty
 \def\@oddfoot{\centerline{\thepage}}%
 \let\@evenfoot\@oddfoot}
\makeatother

\makeatletter
\def\@author#1{\g@addto@macro\elsauthors{\normalsize%
    \def\baselinestretch{1}%
    \upshape\authorsep#1\unskip\textsuperscript{%
      \ifx\@fnmark\@empty\else\unskip\sep\@fnmark\let\sep=,\fi
      \ifx\@corref\@empty\else\unskip\sep\@corref\let\sep=,\fi
      }%
    \def\authorsep{\unskip,\space}%
    \global\let\@fnmark\@empty
    \global\let\@corref\@empty  %% Added
    \global\let\sep\@empty}%
    \@eadauthor={#1}
}
\usepackage{booktabs,diagbox, float, adjustbox}
\usepackage{lineno,hyperref}
\modulolinenumbers[5]
\usepackage{array}
\newcolumntype{C}[1]{>{\centering\let\newline\\\arraybackslash\hspace{0pt}}m{#1}}
\usepackage{numcompress}

\begin{document}

\begin{frontmatter}

\title{Deep-Learning for Classification of Colorectal Polyps on Whole-Slide Images}

\author[biomed,compsci]{Bruno Korbar}
\author[dhmc_footnote]{Andrea M. Olofson}
\author[dhmc_footnote]{Allen P. Miraflor}
\author[dhmc_footnote]{Catherine M. Nicka}
\author[dhmc_footnote]{Matthew A. Suriawinata}
\author[compsci]{Lorenzo Torresani}
\author[dhmc_footnote]{Arief A. Suriawinata}
\author[biomed,compsci,epi]{Saeed Hassanpour\corref{mycorrespondingauthor}}

\address[biomed]{Biomedical Data Science Department, The Geisel School of Medicine at Dartmouth} 
\address[dhmc_footnote]{Department of Pathology and Laboratory Medicine, The Geisel School of Medicine at Dartmouth}
\address[compsci]{Computer Science Department, Dartmouth College}
\address[epi]{Epidemiology Department, The Geisel School of Medicine at Dartmouth}
\cortext[mycorrespondingauthor]{Corresponding author}
\ead{saeed.hassanpour@dartmouth.edu}
%%%
%%% ABSTRACT
%%%

\begin{abstract}
Histopathological characterization of colorectal polyps is an important
principle for determining the risk of colorectal cancer and future rates of surveillance for patients.
This characterization is time-intensive, requires years of specialized
training, and suffers from significant inter-observer and intra-observer variability.
In this work, we built an automatic image-understanding method that can
accurately classify different types of colorectal polyps in whole-slide histology
images to help pathologists with histopathological characterization and
diagnosis of colorectal polyps. The proposed image-understanding method is based on deep-learning techniques,
which rely on numerous levels of abstraction for data representation
and have shown state-of-the-art results for various image analysis tasks.
Our image-understanding method covers all five polyp types (hyperplastic
polyp, sessile serrated polyp, traditional serrated adenoma, tubular adenoma,
and tubulovillous/villous adenoma) that are included in the US multi-society
task force guidelines for colorectal cancer risk assessment and surveillance,
and encompasses the most common occurrences of colorectal polyps.
Our evaluation on 239 independent test samples shows our proposed method can identify
the types of colorectal polyps in whole-slide images with a high efficacy
(accuracy: 93.0\%, precision: 89.7\%, recall: 88.3\%, F1 score: 88.8\%). 
The presented method in this paper can reduce the cognitive burden on pathologists and improve their accuracy and efficiency
in histopathological characterization of colorectal polyps, and in subsequent risk assessment and follow-up recommendations.

\end{abstract}


\end{frontmatter}

% \linenumbers

%introduction
\section{Introduction}
\label{introduction}
At least half of Western adults will have a colorectal polyp in their
lifetime and one-tenth of these polyps will progress to cancer \citep{wong2009observer}.
If colorectal polyps are detected early, they can be removed before they transform to cancer.
While there are multiple screening methods to detect colorectal polyps, colonoscopy has become
the most common screening test in the United States \citep{lieberman2012guidelines}.
In 2012, US multi-society task force on colorectal cancer issued updated guidelines
on colorectal cancer surveillance after colonoscopy screening\textemdash a key principle of which is risk
assessment and follow-up recommendation based on histopathological characterization of the
detected polyps in the baseline colonoscopy. Therefore, detection and histopathological
characterization of colorectal polyps are an important part of colorectal cancer screening,
through which high-risk colorectal polyps are distinguished from low-risk polyps.
The risk of subsequent polyps and colorectal cancer and the timing of follow-up
colonoscopies depend on this characterization \citep{lieberman2012guidelines}; however,
accurate characterization of certain polyp types can be challenging and there
is a large degree of variability for how pathologists characterize and diagnose
these polyps. As an example, sessile serrated polyps can potentially develop more aggressively into colorectal cancer compared to other colorectal polyps, because of the \textit{serrated pathway} in tumorigenesis \citep{leggett2010role}. The serrated pathway is associated with mutations in the \textit{BRAF} or \textit{KRAS} oncogenes, and CpG island methylation, which can lead to the silencing of mismatch repair genes (e.g., \textit{MLH1}) and a more rapid progression to malignancy \citep{vu2011individuals}. Therefore, differentiating sessile serrated polyps from other types of polyps is critical for an appropriate surveillance \citep{biscotti2005assisted}. Histopathological characterization is the \textit{only reliable existing method} for diagnosing sessile serrated polyps, because other screening methods designed to detect pre-malignant lesions (such as fecal blood, fecal DNA, or virtual colonoscopy) are not well suited for differentiating sessile serrated polyps from other polyps \citep{kahi2015does}. However, differentiation between sessile serrated polyps and innocuous hyperplastic polyps is a challenging task for pathologists \citep{vu2011individuals, aptoula2013mitosis, irshad2014methods, veta2015assessment}. This is because sessile serrated polyps, like hyperplastic polyps, often lack the dysplastic nuclear changes that characterize conventional adenomatous polyps, and their histopathological diagnosis is entirely based on morphological features, such as serration, dilatation, and branching. Accurate diagnosis of sessile serrated polyps and their differentiation from hyperplastic polyps is needed to ensure that patients receive appropriate/frequent follow-up surveillance, and to prevent the patients from being over-screened. However, in a recent colorectal cancer study, more than 7,000 patients underwent colonoscopy in 32 centers\textemdash ultimately, a sessile serrated polyp was not diagnosed in multiple centers despite the statistical unlikeliness of this outcome \citep{snover2011update}. This indicates there are still considerable gaps in the performance and education of pathologists regarding histologic features of colorectal polyps and their diagnostic accuracy \citep{abdeljawad2015sessile}.

In the past years, computational methods have been developed to assist pathologists in the analysis of microscopic images \citep{gurcan2009histopathological, madabhushi2016image, naik2007gland}. These image analysis methods primarily focus on basic structural segmentation (e.g., nuclear segmentation) \citep{nakhleh2006error, raab2005clinical, malkin1998comparison} and feature extraction (e.g., orientation, shape, and texture) \citep{gil2002image, boucheron2008object, sertel2009histopathological, doyle2007automated}. In some methods, these extracted or hand-constructed features are used as an input to a standard machine-learning classification framework, such as a support vector machine \citep{rajpoot2004svm, kallenbach2013immunohistochemistry} or a random forest \citep{sims2003image}, for automated tissue classification and disease grading.

In the field of artificial intelligence, deep-learning computational models, which are composed of multiple processing layers, can learn numerous levels of abstraction for data representation \citep{lecun2015deep}. These data abstractions have dramatically improved the state-of-the-art computer vision and visual object recognition applications, and, in some cases, even exceed human performance \citep{he2015delving}. Currently, deep-learning models are successfully utilized in autonomous mobile robots and self-driving cars \citep{farabet2012scene,hadsell2009learning}. The construction of deep-learning models only recently became practical due to large amounts of training data becoming available through the World Wide Web, public data repositories, and new high-performance computational capabilities that are mostly due to the new generation of graphics processing units (GPUs) needed to optimize these models \citep{lecun2015deep}.

Recent work has proven the deep-learning approach to be superior for tasks of classification and segmentation on histology whole-slide images, compared to the previous image processing techniques \citep{xie2015deep,sirinukunwattana2016locality,janowczyk2016deep}. As examples, deep-learning models have been developed to detect metastatic breast cancer \citep{cruz2013deep}, to find mitotically active cells \citep{ertosun2015automated}, to identify basal-cell carcinoma \citep{malon2013classification}, and to grade brain gliomas \citep{wang2014cascaded} using H\&E-stained images. Particularly, Sirinukunwattana et al. \citep{sirinukunwattana2015stochastic} presented a deep-learning approach for nucleus detection and classification in H\&E-stained images of colorectal cancer. This model was based on a standard 8-layer convolutional network \citep{le1990handwritten} to identify the centers of nuclei and classify them in four categories of epithelial, inflammatory, fibroblastic, and miscellaneous. Janowczyk et al. released a survey of the applications of deep learning in pathology, exploring domains such as lymphocyte detection, mitosis detection, invasive ductal carcinoma detection, and lymphoma classification \citep{janowczyk2016deep}. All models in the survey used the convolutional neural network proposed by Krizhevsky et al. \citep{krizhevsky2012imagenet}.

With the recent expansion in the use of whole-slide digital scanners, high-throughput tissue banks, and archiving of digitized histological studies, the field of digital pathology is ripe for development and application of computational models to assist pathologists in the histopathological analysis of microscopic images, disease diagnosis, and management of patients. Considering these recent advancements in computerized image understanding, and the critical need for computational tools to help pathologists with histopathological characterization and diagnosis of colorectal polyps for more efficient and accurate colorectal cancer screening, we propose a novel deep-learning-based approach for this task.

% materials and methods
\section{Materials and Methods}
\label{materials_head}
The whole-slide images require to develop and evaluate our method were collected from patients who underwent colorectal cancer screening at our academic quaternary care center. Our domain expert pathologist collaborators annotated different types of colorectal polyps in these images. We used these annotations as reference standards for training and testing our deep-learning methods for colorectal polyp classification on whole-slide images, as well as for establishing a deep-learning benchmark for this task. Defining a benchmark for deep-learning methods can provide a guideline for future clinical implementations, and can promote thorough understanding of understanding of an architecture as a critical factor in the deep-learning model’s performance.

\subsection{Dataset}
\label{dataset}
\begin{figure}[h!]
\centering
\includegraphics[width=1\textwidth]{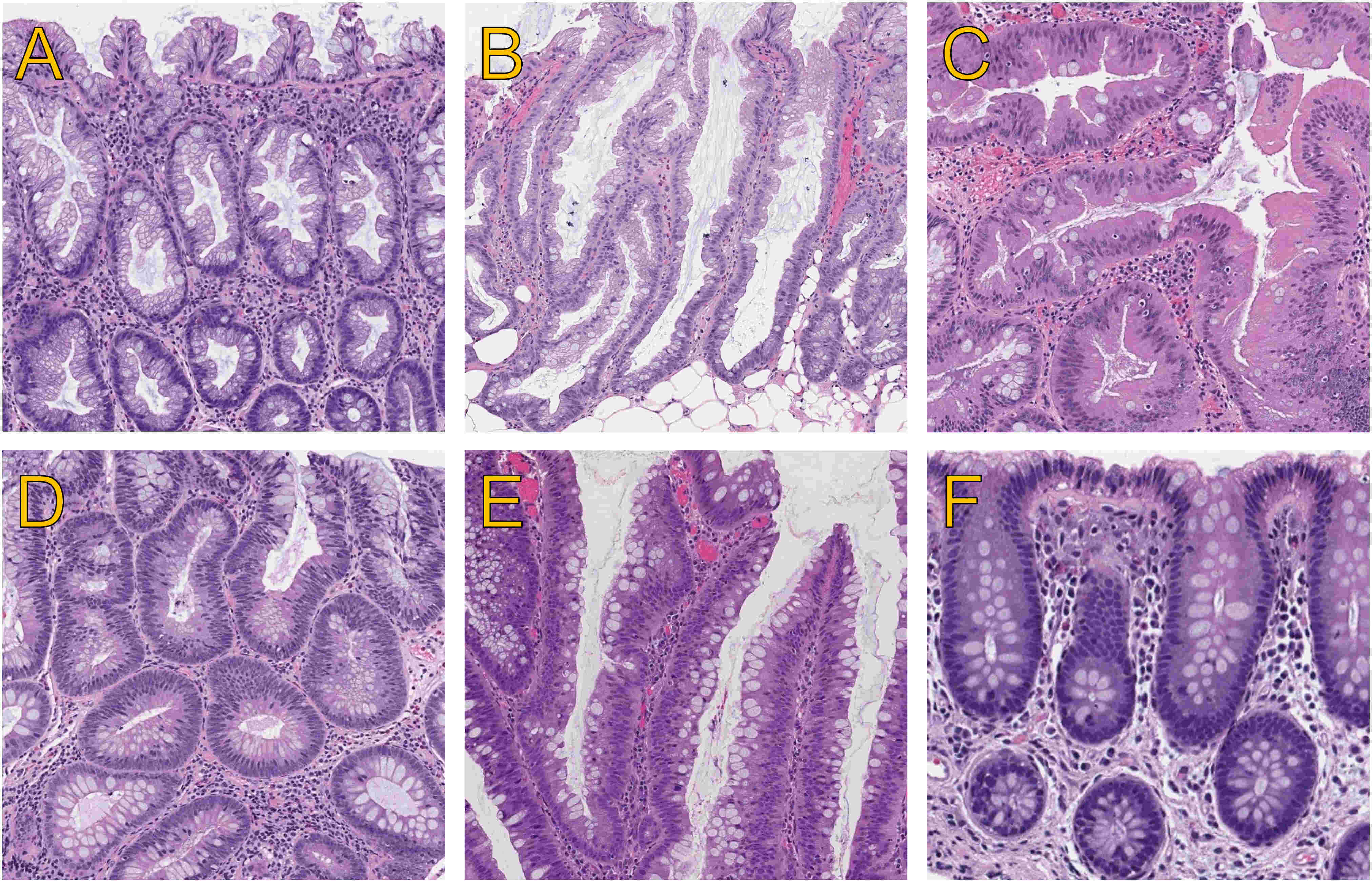}
\caption{H\&E-stained image samples with different histopathological characterizations for colorectal polyps: (A) hyperplastic, (B) sessile serrated, (C) traditional serrated, (D) tubular, (E) tubulovillous/villous, and (F) normal.}
\label{fig:data}
\end{figure}

The data required for training and evaluating the proposed approach in this project is collected from Dartmouth-Hitchcock Medical Center (DHMC) patients who underwent colorectal cancer screening since 01/2010. The Department of Pathology and Laboratory Medicine at DHMC has instituted routine whole-slide scanning for slide archiving, employing three high-throughput Leica Aperio whole-slide scanners. These slides are digitized at $200\times$ magnification. Our histology imaging dataset includes H\&E-stained, whole-slide images for five types of colorectal polyps: hyperplastic polyp, sessile serrated polyp, traditional serrated adenoma, tubular adenoma, and tubulovillous/villous adenoma. These five classes cover the most common occurrences of colorectal polyps, and encompass all polyp types that are included in the US multi-society task force guidelines for colorectal cancer risk assessment and surveillance \citep{lieberman2012guidelines}. In addition, our dataset will include normal samples, which do not contain colorectal polyps, for our model training and evaluation. Figure \ref{fig:data} shows sample H\&E-stained images from all colorectal polyp types that were collected in this project.

For this project, 1,723 whole-slide images have been collected through this collaboration with the Department of Pathology and Laboratory Medicine at DHMC. The number of collected images from each colorectal polyp type is presented in Table 2. We used 85\% of the collected images in this dataset, as the training set, and evaluated its performance on the remaining 15\% as the validation set. An additional 239 whole-slide images were collected after the training for final evaluation. The use of these data for this project is approved by the Dartmouth Institutional Review Board. 

\subsection{Image Annotation}
\label{annotation}
High-resolution histology images for colorectal polyp samples are large---most of the slides encompass normal tissue and only a small part of a whole-slide image is actually related to the colorectal polyp. In this study, two collaborators, resident pathologists from the Department of Pathology and Laboratory medicine at DHMC, independently reviewed the whole-slide images in our training and test sets to identify the type of colorectal polyps in images, as reference standards. In addition, to train a classification model on colorectal polyp features in these slides, and as a preprocessing step, one of the pathologists outlined the regions in which the colorectal polyp was present and generated smaller crops focused on colorectal polyps. Extracting smaller crops for training deep-learning classifiers has shown superior performance in previous histopathology analysis applications \citep{bengio2009learning}. A second, highly experienced pathologist also reviewed the whole-slide images and their associated extracted crops. The disagreements in classifying and cropping the images were resolved through further discussions between the annotator pathologists and through consultation with a third, senior gastrointestinal pathologist collaborator. To ensure the accuracy of these manual annotations and resulting image crops, when an agreement could not be reached on a polyp type or cropping for an image, that image was discarded and replaced by a new image.

\subsection{Training Architecture and Framework}
\label{architecture}
Deep-learning is strongly rooted in previously existing artificial neural networks \citep{lecun2015deep}, although the construction of deep-learning models only recently became practical due to the availability of large amounts of training data and new high-performance GPU computational capabilities designed to optimize these models \citep{lecun2015deep}. Krizhevsky et. al. developed a deep learning model \citep{krizhevsky2012imagenet} based on convolutional neural networks (ConvNets) \citep{le1990handwritten} that significantly improved the image classification results and reduced the error rate about 10\% compared to the best non-deep-learning methods’ performance in computer vision at the time. Since then, various deep-learning methods have been developed and have improved the models' performance even further.

While it has been shown that the increase in depth would yield superior results \citep{simonyan2014very}, the state-of-the-art deep-learning models were unable to take advantage of this increase, beyond 50 layers \citep{szegedy2015going,simonyan2014very}. This was because of a fundamental problem with propagating gradients for optimizing networks with large number of layers, which is commonly known as the \textit{vanishing gradient problem} \citep{simonyan2013deep,he2015deep}. Therein, beyond a moderate number of layers, the models experience performance degradation according to the degree of increase in the number of layers in previous architectures. In 2015 Microsoft introduced “residual architecture” (ResNet), which addressed the vanishing gradient problem. Upon its introduction, ResNet outperformed previous architectures by significant margins in all main tracks of the ImageNet computer vision competition, including object detection, classification, and localization \citep{he2015deep}, and allowed for up to 152 layers before experiencing the performance degradation. To empirically support our choice of architecture, we conducted an ablation study on top performing deep-learning architectures \citep{russakovsky2015imagenet}, such as AlexNet \citep{krizhevsky2012imagenet}, VGG \citep{simonyan2014very}, GoogleNet \citep{szegedy2015going}, and different variations of ResNet \citep{he2015deep}. Results of this comparison can be found in \hyperref[table:ablation]{Table 1} in the Results section.

\begin{figure}[h!]
\centering
\includegraphics[width=0.9\textwidth]{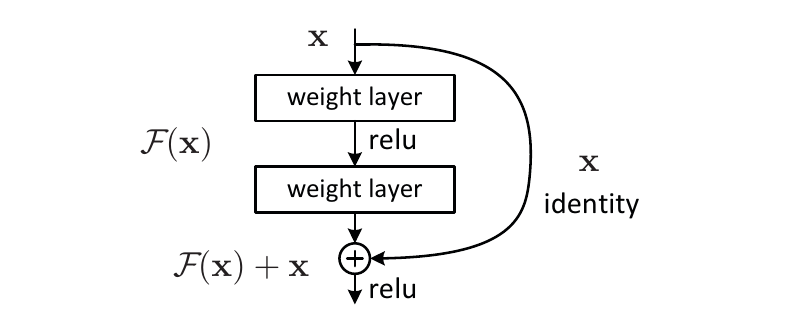}
\caption{The mechanism of a sample residual black in the ResNet architecture \citep{he2015deep}}
\label{fig:block}
\end{figure}

For our approach, we have adopted a modified version of a residual architecture, as this approach yielded state-of-the art performance in both image recognition benchmarks, ImageNet \citep{russakovsky2015imagenet}, COCO \citep{tsung2014coco}, as well as in image segmentation benchmarks, COCO-segmentation \citep{tsung2014coco}. We implement ResNet as a standard neural network, consisting of $3\times 3$ and $1 \times 1$ convolution filters, and introduced additional mappings or shortcuts that bypass several convolutional layers. Inputs from these additional mappings were then added with the output of the previous layer to form a residual input to the next layer such as in Figure \ref{fig:block}. Introduction of these shortcuts almost completely eliminates the vanishing gradient problem, which in term allows for greater depth of the neural networks while keeping the computational complexity at a manageable level due to relatively small convolutional filters. In addition to the identity mappings, we experimented with “projection shortcuts” (done by $1\times 1$ convolution) when dimensions of the shortcuts did not match the dimensions of the preceding layer in order to achieve the best performance in our study \citep{he2015deep}.

\subsection{Training}
\label{training}
To verify our architecture choice in this work, we further separated 15\% of the formerly mentioned training data as the hold-out validation set to run an ablation study on various deep-learning architectures. After finding the optimal architecture on this validation set, training was repeated on the entire augmented training set. Finally, we evaluated the trained model on our test set. 

Our deep-learning classification model is trained for detecting colorectal polyps in small patches of H\&E-stained, whole-slide images. Each crop is processed as follows.

We first rescale the data to conform to the median of the dimensions along x and y axes computed on a random subset of images. This random subset was confined to 15\% of our training set for computational efficiency. If the image size along any dimensions was below the median, we use zero-padding to make it conform to the aspect ratio. We normalize each image using mean and standard deviation computed on training data in order to neutralize color differences caused by inconsistent staining of the slides. For color jittering data augmentation, we compute PCA for all points of a subset of training data, sample the offset along principal components, and add it to all pixels of each image. Finally, we rotate each image by 90 degrees to enforce rotational invariance, and flip a randomly-selected 50\% of the images along the horizontal axis.

\begin{figure}[ht!]
\centering
\includegraphics[width=0.8\textwidth]{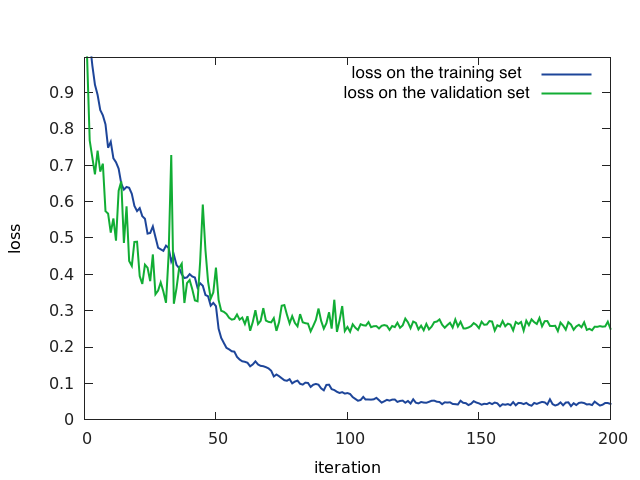}
\caption{Training loss per iteration for 152 layer ResNet model on training and validation sets.}
\label{fig:loss}
\end{figure}

We trained the optimal model for 200 epochs on the augmented training set, with initial learning rate of 0.1, decreasing it 0.1 times each 50 epochs, and 0.9 momentum. Overall training time for different architectures took 36 hours on a single NVIDIA K40c GPU. Figure \ref{fig:loss} shows the value of the loss function on the training and validation sets for training a ResNet model with 152 layers. As can be seen in this figure, the model converges early in the training process near 50th epoch.

\subsection{Inferencing Classes for Whole-Slide Images}
\label{inference}
\begin{figure}[h!]
\centering
\includegraphics[width=1\textwidth]{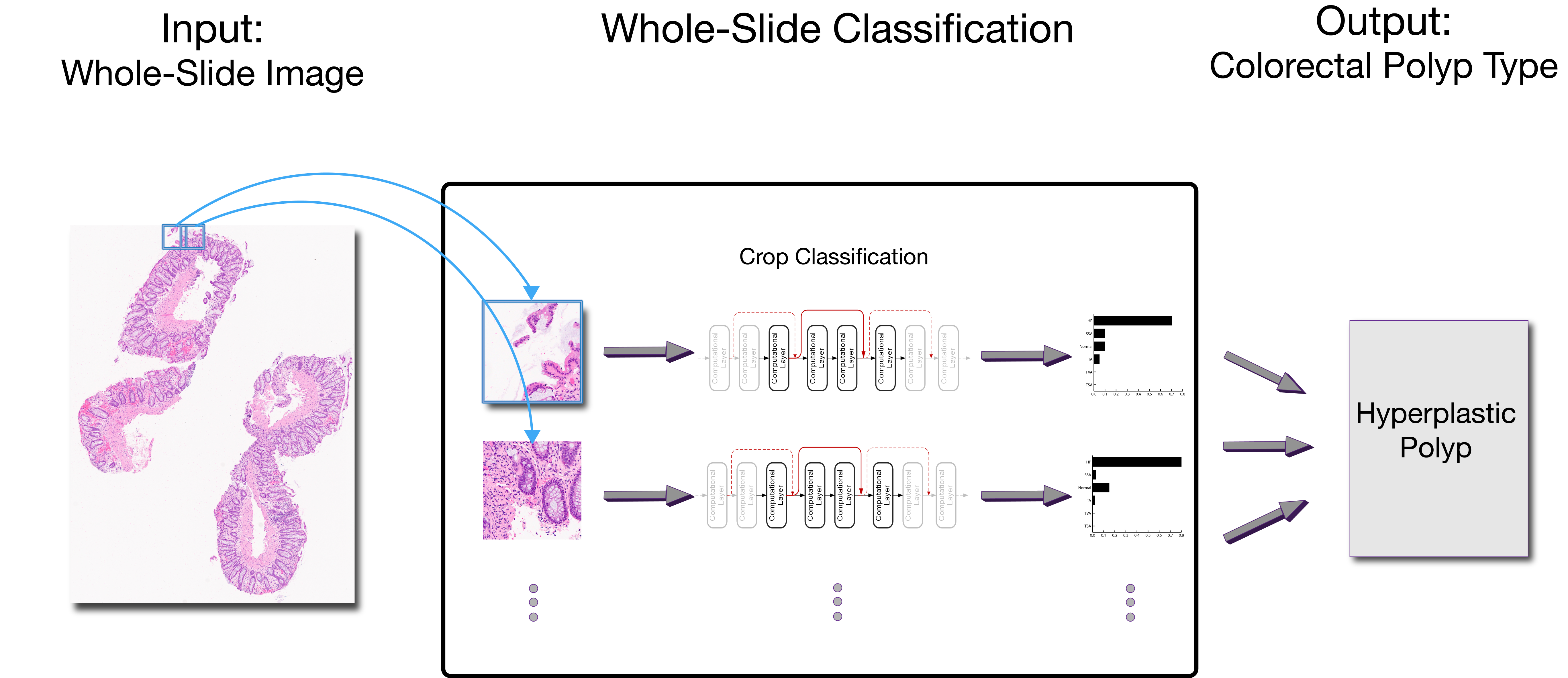}
\caption{Overview of our approach for classification of colorectal polyps in whole-slide, H\&E-stained images.}
\label{fig:overview}
\end{figure}

As mentioned in the Training section, our deep-learning classification model is trained for detecting colorectal polyps in small patches of H\&E-stained, whole-slide images. To identify the colorectal polyps and their types in whole-slide images by our deep-learning model, we break the whole-slide images into smaller, overlapping patches and apply the model on these patches. \hyperref[fig:overview]{Figure 4} shows the overview of our approach for whole-slide image classification. In this work, we use overlapping patches enforcing one-third (i.e. 33\%) overlap to cover the full image. In order to extract coherent patches with image crops used for training, the size of these patches is fixed at the median size of a random 15\% subset of the image crops from our training set. Our system infers the type of colorectal polyp in the whole-slide image based on the most common colorectal polyp class among the associated patches for a whole-slide image. In addition, to reduce the noise and increase the confidence of our results, we only associate a class to a whole-slide image if at least a minimum of 5 patches are identified as that class, with 70\% average confidence. If there is no support for any of the colorectal polyp types among the patches, the whole-slide image is classified as “normal”.
\vspace*{0.5cm}

\subsection{Evaluation}
\label{evaluation}
At training time, we evaluated our models using a validation set of images cropped as described in the section \ref{dataset}. Based on these results we could evaluate the per-crop accuracy in order to understand and address potential pitfalls and inter-class confusion. For the evaluation of the final model, we applied our proposed inference mechanism on whole-slide images in the test set. In this evaluation, we measure the standard machine-learning evaluation metrics of accuracy, sensitivity (recall), specificity, positive predictive value (precision), negative predictive value, and F1 score for our method \citep{powers2011evaluation}. In addition, we calculate 95\% confidence intervals for all of the performance metrics in this evaluation through the Clopper-Pearson method \citep{clopper1934use}.

% Results
\section{Results}

\begin{table}[H]
\begin{adjustbox}{center}

\begin{tabular}{ |C{3.5cm}||C{2cm}|C{2cm}|C{3cm}|C{3cm}|  }
\hline
\multicolumn{5}{|p{14.5cm}|}{
\textbf{Table 1:} Results of ablation test on raw image crops over 50 epochs for selecting the best deep-neural network architecture.
} \\
\hline
\textbf{Architecture}&
\textbf{Number of layers}&
\textbf{Accuracy}&
\textbf{95\% confidence interval}&
\textbf{Evaluation time in seconds}
\\
\hline
AlexNet\citep{krizhevsky2012imagenet} & 8&
71.8\% & (65.4\% - 77.6\%)& 2.5 \\
 VGG\citep{simonyan2013deep} & 19 &
76.4\% & (70.2\% - 81.8\%) & 3.0 \\
GoogleNet \citep{szegedy2015going} & 22 &
88.7\%& (83.8\% - 92.5\%)& 2.4\\
ResNet-A \citep{he2015deep}& 50 &
81.2\%& (75.4\% - 86.1\%)& 2.2\\
ResNet-B \citep{he2015deep}& 101 &
82.7\%& (77.1\% - 87.4\%)& 2.6\\
ResNet-C \footnotemark[1] \citep{he2015deep}& 152 &
87.1\% & (82.0\% - 91.2\%) & 3.1\\
ResNet-D \footnotemark[2] \citep{he2015deep}& 152 &
89.0\% & (84.1\% - 92.8\%) & 3.1\\
\hline
\end{tabular}

%\caption{Ablation results}
\label{table:ablation}
\end{adjustbox}

\end{table}

\footnotetext[1]{152 layer ResNet with identity mappings}
\footnotetext[2]{152 layer ResNet with projection mappings}
%%%%%%%%%%%%%%%%%%%%

\begin{table}[H]
\begin{adjustbox}{center}

\label{table:byclass_crops}
\begin{tabular}{ |C{5cm}||C{2cm}|C{2cm}|C{3cm}|  }
\hline
\multicolumn{4}{|p{13cm}|}{
\textbf{Table 2:} Results of our best model (ResNet-D) for classification of colorectal polyps in cropped histology images based on validation data.}
\\
\hline
\textbf{Colorectal polyp type}&
\textbf{\# Cases in the test set}&
\textbf{Accuracy}&
\textbf{95\% confidence interval}
\\
\hline

Hyperplastic polyp&
34& 86.9\% & (81.5\% - 91.3\%)\\
Sessile serrated polyp &
33 & 87.4\% & (82.0\% - 91.7\%) \\
Traditional serrated adenoma &
38& 91.5\%& (86.7\% - 94.9\%) \\
Tubular adenoma&
35& 94.5\%& (90.4\% - 97.2\%)\\
Tubulovillous/villous adenoma&
29& 91.5\%& (86.7\% - 94.9\%)\\
Normal&
30& 96.0\%& (92.3\% - 98.2\%)\\
\hline
\textit{Total}& 
\textit{199}& \textit{91.3\%}& \textit{(86.5\% - 94.8\%)}\\
\hline

\end{tabular}
\end{adjustbox}

\end{table}

%%%%%%%%%

\begin{table}[H]
\begin{adjustbox}{center}

\label{table:byclass_WHI}
\begin{tabular}{ |C{2cm}||C{1.9cm} | C{1.9cm} |C{1.9cm}|C{1.9cm}|C{1.9cm}|C{1.9cm}|C{2cm}|  }
\hline
\multicolumn{8}{|p{18cm}|}{
\textbf{Table 3:} Results of our final model for classification of colorectal polyps in 239 whole-slide images in our test set (HP: hyperplastic polyp, SSP: sessile serrated polyp, TSA: traditional serrated adenoma, TA: tubular adenoma, and TVA/V: tubulovillous/villous adenoma).}
\\
\hline
& \textbf{HP} \hspace{1cm} (N = 37)& \textbf{SSP}\hspace{1cm} (N = 39)& \textbf{TSA} (N=38) & \textbf{TA} \hspace{1cm} (N=39) & \textbf{TVA/V} (N=38)& \textbf{Normal} (N=48)& \textit{\textbf{Total} (N=239)}\\
\hline

\textbf{Accuracy}&
89.8\%  \footnotesize(85.3\%-93.3\%) & 
89.5\% \footnotesize (85.0\%-93.1\%)&
94.7\% \footnotesize(91.1\%-97.2\%)&
93.1\% \footnotesize(89.2\%-96.0\%) &
95.8\% \footnotesize(92.5\%-97.9\%) &
95.0\% \footnotesize(91.5\%-97.4\%) &
\textit{93.0\% \footnotesize(89.0\%-95.9\%)} \\

\textbf{Precision}&
90.9\%
\footnotesize (86.6\%-94.2\%) &
86.11\% \footnotesize(81.1\%-90.2\%)&
100.0\% \footnotesize(98.5\%-100.0\%)&
83.3\% \footnotesize(78.0\%-87.8\%) &
97.2\% \footnotesize(94.3\%-98.9\%)&
80.7\% \footnotesize(75.1\%-85.5\%)&
\textit{89.7\% \footnotesize(85.2\%-93.2\%)} \\

\textbf{Recall}&
81.1\% \footnotesize(75.5\%-85.8\%)&
81.6\% \footnotesize(76.1\%-86.3\%)&
89.5\% \footnotesize(84.9\%-93.0\%)&
89.7\% \footnotesize(85.2\%-93.3\%)&
92.1\% \footnotesize(88.0\%-95.2\%)&
95.8\% \footnotesize(92.5\%-98.0\%)&
\textit{88.3\% \footnotesize(83.6\%-92.1\%)} \\

\textbf{F1 Score}&
85.7\% \footnotesize(80.6\%-89.9\%)&
83.8\% \footnotesize(78.5\%-88.2\%)&
94.4\% \footnotesize(90.8\%-97.0\%)&
86.4\% \footnotesize(81.4\%-90.5\%)&
94.6\% \footnotesize(90.9\%-97.1\%)&
87.6\% \footnotesize(82.8\%-91.5\%)&
\textit{88.8\% \footnotesize(84.1\%-92.5\%)}\\

\hline

\end{tabular}
\end{adjustbox}

\end{table}

%%%%%%%%%%%

\begin{table}[H]
\begin{adjustbox}{center}

\label{table:conf}
\begin{tabular}{ |C{2.5cm}|C{1.5cm} |C{1.5cm}|C{1.5cm}|C{1.5cm}|C{1.5cm}|C{1.5cm}|  }
\hline
\multicolumn{7}{|p{13.9cm}|}{
\textbf{Table 4:} Confusion matrix of our final model for classification of colorectal polyps in 239 whole-slide images on our test set (HP: hyperplastic polyp, SSP: sessile serrated polyp, TSA: traditional serrated adenoma, TA: tubular adenoma, and TVA/V: tubulovillous/villous adenoma).}
\\
\hline
\diagbox[width=8em]%
{Prediction}{Reference} & HP&SSP&TSA&TA&TVA/V&Normal\\
\hline 
HP&30&3&0&0&0&0\\
\hline
SSP
&5
&31
&0
&0
&0
&0\\
\hline
TSA&
0&
0&
34&
0&
0&
0\\
\hline
TA&
0&
0&
2&
35&
3&
2\\
\hline
TVA/V&
0&
0&
0&
1&
35&
0\\
\hline
Normal&
2&
4&
2&
3&
0&
46\\
\hline
\end{tabular}
\end{adjustbox}
\end{table}

% Discussion
\section{Discussion}
\label{discussion}
In this work, we presented an automated system to facilitate the histopathological characterization of colorectal polyps on H\&E-stained, whole-slide images with high sensitivity and specificity. Our evaluation shows that our system can accurately differentiate high-risk polyps from both low-risk colorectal polyps and normal cases by identifying the corresponding colorectal polyp types, such as hyperplastic, sessile serrated, traditional serrated, tubular, and tubulovillous/villous, on H\&E-stained, whole-slide images. These polyp types are the focus of and major criteria in the US multi-society task force guidelines for colorectal cancer surveillance and cover most colorectal polyp occurrences \citep{lieberman2012guidelines}. This project is inspired in part by the use of image analysis software in Papanicolaou (Pap) smear screening \citep{biscotti2005assisted} for cervical cancer. In past years, the automation of Pap smear screening has dramatically improved the diagnostic accuracy and screening productivity, and helped to reduce the incidence of cervical cancer and mortality among American women \citep{biscotti2005assisted}. Our proposed system can potentially achieve a similar impact on colorectal cancer screening, as colorectal cancer is the second leading cause of cancer death among both men and women in the United States \citep{americancanc2014stats}, and colorectal polyps are the most common findings during colorectal cancer screening \citep{lieberman2012guidelines}.

Our proposed automatic image understanding system can potentially reduce the time needed for screening analysis, diagnosis, and prognosis; reduce the manual burden on clinicians and pathologists; and significantly reduce the potential errors arising from the histopathological characterization of colorectal polyps for the subsequent risk assessment and follow-up recommendations. By combining the outcomes of our proposed system with pathologists' interpretations, this technology will be able to significantly improve the accuracy of diagnoses and prognoses, and therefore foster precision medicine. Along those lines, this project will provide a platform for improved quality assurance of colorectal cancer screening and understanding of common error patterns to improve clinical training. In the clinical setting, the implementation of our approach will enhance the accuracy of colorectal cancer screening, reduce the cognitive burden on pathologists, positively impact patient health outcomes, and reduce colorectal cancer mortality by fostering early preventive measures. Improvement in the efficiency of colorectal cancer screening will result in a reduction in screening costs, an increase in the coverage of screening programs, and an overall improvement in public health.

This project leverages ResNet architecture \citep{he2015deep}, a new deep-learning paradigm, to address the “vanishing gradient” problem in model training. This architecture enables the development of ultra-deep models with superior accuracy for characterization of histology images in comparison to existing approaches. Our ablation test results confirm (\hyperref[table:ablation]{Table 1}) the superiority of ResNet deep-neural network architecture with 152 layers for our classification task in comparison to other common architectures such as AlexNet \citep{krizhevsky2012imagenet}, VGG \citep{simonyan2013deep}, and GoogleNet \citep{szegedy2015going}. Although this best performing ResNet model has significantly more layers than other architectures in this comparison, its evaluation time (3.1 seconds) is close to the other models in a practical range. This small evaluation time difference is due to relatively simple computational layers in ResNet architecture. In addition, as can be seen in  \hyperref[table:byclass_crops]{Table 2}, data augmentation has a positive impact on the accuracy of our classification results.

We evaluated our ResNet-based, whole-slide inference model for colorectal polyp classification on 239 independent whole-slide, H\&E-stained images. These results are presented in Tables \ref{table:byclass_WHI} and \hyperref[table:conf]{4}. As we can see in these tables, our whole-slide inferencing approach demonstrates a strong performance across different classes, with an over all accuracy of 93.0\%, an over all precision of 89.7\%, an over all recall of 88.3\%, and an over all F1 score of 88.8\%. As can be seen in the presented confusion matrix (\hyperref[table:conf]{Table 4}), in this evaluation we observed a tendency to classify low-confidence examples as normal. This may be due to the diversity of whole-slide images that are considered to be normal in our training set. Furthermore, we can see that differentiation between hyperplastic polyps and sessile serrated polyp is another major source of mistakes for our model, which is aligned with gastrointestinal pathologists' experience in this task \citep{vu2011individuals, irshad2014methods, veta2015assessment}.

Although our proposed histopathology characterization system is based on strong deep-learning methodology, and achieved a strong performance in our evaluation on the test set collected at our organization, we still plan to take additional steps to improve our evaluation and results. One possible improvement could be a further increase in our architecture's number of layers, which requires collecting a larger training set. To this end, through a collaboration with the New Hampshire Colonoscopy Registry (NHCR), we are planning to apply and evaluate the proposed method on an additional dataset from patients across New Hampshire for the external validation of our approach.

One shortcoming of our system for histopathological characterization, and deep-learning models in general, is the “black box” approach to the outcomes. These image analysis methods are mostly focused on the efficacy of the final results and rarely provide sufficient evidence and details on factors that contribute to their outcomes. As future work, we aim to leverage visualization methods for deep learning models to tackle this problem. These visualization methods will provide insight about influential regions and features of a whole-slide image that contribute to the histopathological characterization results. This visualization will help pathologists verify the characterization results of our method and understand the underlying reasoning for a specific classification.

Our proposed method to characterize colorectal polyps in whole-slide images can be extended to other histopathology analyses and prognosis assessment problems outside of colorectal cancer screening. The proposed method for whole-slide, H\&E-stained histopathology analysis builds an illustrative “showcase” for colorectal cancer screening. As future work, we plan to build training sets for other challenging histopathology characterization problems and extend the developed deep-learning image analysis framework to histopathology image analysis and assessment in other types of cancer, such as melanoma, glioma/glioblastoma, and breast carcinoma.

% Conclusion
\section{Conclusion}
\label{conclusion}
In this paper, we presented an image understanding system to assist pathologists in characterization of colorectal polyps on H\&E-stained, whole-slide images. This system was based on state-of-the-art, deep-neural network architecture to identify the types of colorectal polyps in whole-slide, H\&E-stained images. We evaluated our developed system on 239 H\&E-stained, whole-slide images for detection of five colorectal polyp classes outlined by the US multi-society task force guidelines for colorectal cancer risk assessment and surveillance. Our results (Accuracy: 93.0\%, Precision: 89.7\%, Recall: 88.3\%, F1 Score: 88.8\%) show the efficacy of our approach for this task. The technology developed and tested in this work has a great potential to be highly impactful by serving as a low-burden, efficient, and accurate diagnosis and assessment tool for colorectal polyps. Therefore, the outcomes of this project can potentially increase the coverage and accuracy of colorectal cancer screening programs, and overall reduce colorectal cancer mortality.

\section{Acknowledgments}
We would like to thank Haris Baig and Du Tran from Visual Learning Group at Dartmouth College for helpful discussions.

\section*{References}
\bibliography{mybibfile}

\end{document}